\documentclass[letterpaper, 10 pt, conference]{ieeeconf}  % Comment this line out if you need a4paper

\IEEEoverridecommandlockouts                              % This command is only needed if 
\usepackage{algorithmic}
\usepackage{amsmath, amssymb, amsfonts}
\usepackage{graphics} % for pdf, bitmapped graphics files
\usepackage{graphicx} %
\usepackage{stfloats}
\usepackage{float} %
\usepackage{subfigure} %
\usepackage{threeparttable}
\usepackage{multirow}
\usepackage{array}
\usepackage{algorithm}
\usepackage{algorithmic}
\usepackage[normalem]{ulem}
\usepackage[colorlinks,linkcolor=red]{hyperref}
\usepackage{cite}
\usepackage{amsmath}

\title{\LARGE \bf
Rethinking 6-Dof Grasp Detection: A Flexible Framework for High-Quality Grasping
}

\author{Pengwei Xie$^{1,\dagger}$, Siang Chen$^{1,3,\dagger}$, Wei Tang$^{2,\dagger}$, Dingchang Hu$^{1}$, Wenming Yang$^{2}$, Guijin Wang$^{1,3,*}$%
\thanks{$^{1}$The Department of Electronic Engineering, Tsinghua University, Beijing 100084, China.}
\thanks{$^{2}$The Department of Electronic Engineering, Shenzhen International Graduate School, Tsinghua University, Shenzhen 518071, China.}
\thanks{$^{3}$Shanghai AI Laboratory, Shanghai 200232, China.}
\thanks{$^{\dagger}$Equal Contribution.}
\thanks{*Correspondence: {\tt wangguijin@tsinghua.edu.cn}}
\thanks{This work was partly supported by the Special Foundations for the Development of Strategic Emerging Industries of Shenzhen(Nos.CJGJZD20210408092804011\&JSGG20211108092812020).}
}
\begin{document}

\maketitle
\thispagestyle{empty}
\pagestyle{empty}

% 
%%%%%%%%%%%%%%%%%%%%%%%%%%%%%%%%%%%%%%%%%%%%%%%%%%%%%%%%%%%%%%%%%%%%%%%%%%%%%%%%
\begin{abstract}
Robotic grasping is a primitive skill for complex tasks and is fundamental to intelligence. For general 6-Dof grasping, most previous methods directly extract scene-level semantic or geometric information, while few of them consider the suitability for various downstream applications, such as target-oriented grasping. Addressing this issue, we rethink 6-Dof grasp detection from a grasp-centric view and propose a versatile grasp framework capable of handling both scene-level and target-oriented grasping. Our framework, \textit{FlexLoG}, is composed of a \textit{Flex}ible Guidance Module and a \textit{Lo}cal \textit{G}rasp Model. Specifically, the Flexible Guidance Module is compatible with both global (e.g., grasp heatmap) and local (e.g., visual grounding) guidance, enabling the generation of high-quality grasps across various tasks. The Local Grasp Model focuses on object-agnostic regional points and predicts grasps locally and intently. Experiment results reveal that our framework achieves over 18\% and 23\% improvement on unseen splits of the GraspNet-1Billion Dataset. Furthermore, real-world robotic tests in three distinct settings yield a 95\% success rate.

\end{abstract}

%%%%%%%%%%%%%%%%%%%%%%%%%%%%%%%%%%%%%%%%%%%%%%%%%%%%%%%%%%%https://www.overleaf.com/project/652a5aeeab8c86e3cd87402c%%%%%%%%%%%%%%%%%%%%%
\section{INTRODUCTION}

Robotic grasping is fundamental to various complex robotic manipulation tasks across various fields, including manufacturing, service industries, and medical assistance. Despite its critical importance, the efficiency and quality of grasp detection across diverse scenarios remain unsatisfactory, particularly for target-oriented grasping. 
% Data-driven methods have developed rapidly in recent years due to their generalization ability to unknown objects. 

% Object grasping is a critical component of robotics in manufacturing, service, medical assistance, etc. Despite its vital importance, fast and accurate grasping is still challenging for robots. Recent advances in deep learning have enabled data-driven methods to generalize to unseen objects. 

Recent advances in deep learning have enabled data-driven methods for generating grasps of objects without 3D models. Representative methods \cite{morrison2018closing,kumra2020antipodal} generate planar grasps similar to rotated object detection and achieve good performance in simple scenarios with high efficiency. However, this representation inherently constrains the gripper to be perpendicular to the camera plane, limiting its applicability in specific contexts.

\begin{figure}[ht]
% \vspace{-0.8cm} % adjust distance above distance 
\centering
    {\includegraphics[width=9cm]{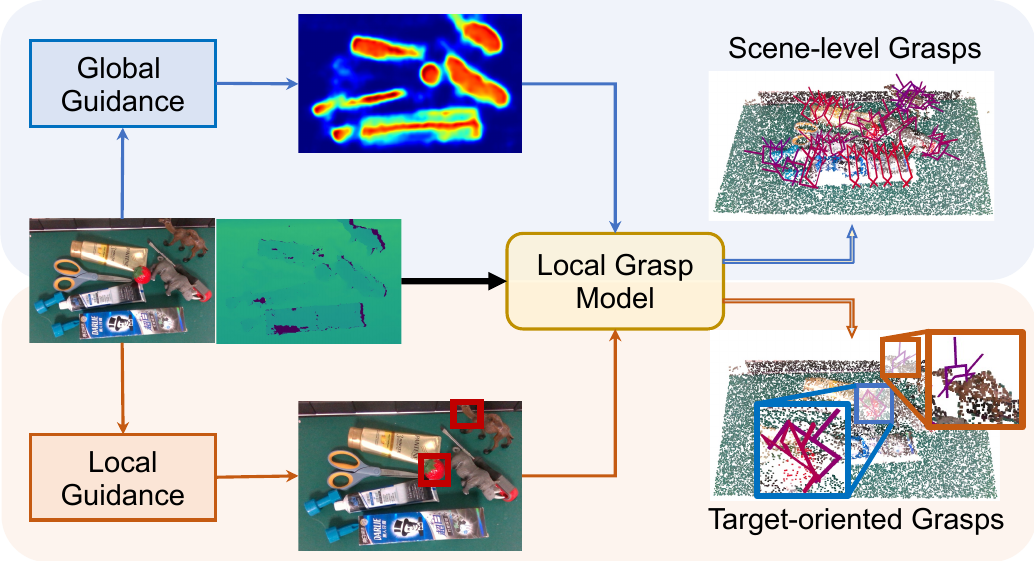}}
        \caption{Our framework can be flexibly integrated with global or local guidance methods for different application scenarios. Global guidance, such as grasp heatmap, can be utilized to generate scene-level grasps. Local guidance, such as object detection, can be utilized to generate target-oriented grasps.}
    \label{fig:teaser} 
\end{figure}

Accordingly, 6-Dof grasping has drawn extensive attention because of its broader applications. Pioneer methods employ a sampling-evaluation strategy \cite{ten2017grasp,liang2019pointnetgpd} or a direct regression paradigm \cite{ni2020pointnet++, qin2020s4g} to predict grasp attributes, which are either time-consuming or inaccurate. Advanced works \cite{zhao2021regnet,wei2021gpr,wang2021graspness,chen2023hggd} excavate local features based on global information, implicitly split the grasp generation into two stages: \textit{\textbf{where}} and \textit{\textbf{how}} \cite{wang2021graspness}. The \textit{\textbf{where}} stage encodes the global information to provide scene-level guidance helping to locate grasps, and then the \textit{\textbf{how}} stage generates refined grasps based on it. Despite the impressive performance, these approaches are flawed in two ways. Firstly, they are limited to integrating only with scene-level guidance in their \textit{\textbf{where}} stage, without considering the suitability for different downstream applications, such as target-oriented grasping aiming to grasp specific objects or parts. For target-oriented grasping, \cite{murali20206,sundermeyer2021contact,liu2022ge} first generate scene-level grasps and apply grasp filtering. These approaches introduce unnecessary computational load and risk interference from irrelevant objects, potentially leading to low-quality grasps or even no grasp in the targeted area. Secondly, the quality of their generated grasps is still unsatisfying, especially for unseen objects. 

Different from these scene-level methods, we rethink the 6-Dof grasp detection problem in a grasp-centric view and turn our attention from the scene-level to the region-level. We propose a novel flexible framework capable of handling both scene-level and target-oriented grasping. Concretely, our framework, named \textbf{\textit{FlexLoG}}, is composed of a \textbf{\textit{Flex}}ible Guidance Module and a \textbf{\textit{Lo}}cal \textbf{\textit{G}}rasp Model. The Flexible Guidance Module is compatible with both global and local guidance, aggregating multiple local regions. Subsequently, the Local Grasp Model processes the regional data points and efficiently predicts grasps within each region.

% Despite the impressive performance, these approaches are flawed in three ways. Firstly, few of them consider their suitability for different downstream applications. For example, if the task is grasping a specific object or object part, these approaches \cite{murali20206} require generating grasps of the whole scene, bringing unnecessary computing overhead and interference from unconcerned objects. Secondly, the quality of their generated grasp is still unsatisfying, especially for unseen objects. Thirdly, they generalize poorly to new scenarios, such as drastic changes in camera viewpoints or beyond the table-top case.

% In this work, we turn our attention from scene-level to part-level, constructing a novel framework that is explicitly split into two stages: \textit{\textbf{where}} and \textit{\textbf{how}}. For the \textit{\textbf{how stage}}, we design an effective and efficient network that takes part-centric data and predicts grasps at part-level. 
% In this letter, we focus on the \textit{\textbf{how to grasp}} stage, constructing a novel framework to generate grasps efficiently and locally. To cooperate with it, a simple but robust Gaussian-based approach is proposed to generate millions of local regional points with the local grasp annotations. As for the \textit{\textbf{where stage}}, we develop a coarse-to-fine heuristic strategy, which detect and augment high-quality grasps in a local-to-global way. 

There are several highlights to reformulate this problem in a local grasp-centric view. Firstly, as shown in Fig. \ref{fig:teaser}, our framework can be guided by either global or local guidance methods (e.g., scene-level heatmap\cite{chen2023hggd}, object-level detection\cite{ren2015faster}), and generates high-quality scene-level grasps or target-oriented grasps efficiently according to demands. Secondly, the grasp-centric representation makes the network capable of learning grasp-related and object-agnostic geometric features, resulting in a sizeable 23\% improvement for novel objects of GraspNet-1Billion \cite{fang2020graspnet}. Additionally, our pipeline exhibits greater flexibility and superior generalization capabilities in novel scenarios compared to other scene-level approaches, making it more adaptable and effective in a broader range of applications. Finally, we deploy our method to the robot and conduct real-world experiments under three distinct settings: Cluttered, Randomly Arranged, and Click-and-Grasp, achieving an impressive average success rate of 95\%.

% In summary, our primary contributions are as follows: 
% \begin{itemize}

%     \item We introduce a flexible and efficient 6-Dof grasp detection framework capable of handling both scene-level and target-oriented grasping.

%     \item We reformulate the 6-Dof grasp detection problem from scene-level to region-level in a grasp-centric view, achieving much performance improvement, particularly for unseen objects.

%     \item We conducted real-world experiments under three distinct settings: Cluttered, Randomly Arranged, and Click-and-Grasp, achieving an impressive average success rate of 95\%.

    % \item Our framework is flexible for various guidance methods for different downstream tasks, provided as a foundation robotic grasping model.

% \end{itemize}

\section{RELATED WORKS}

\subsection{Scene-level Grasping}

Scene-level grasping means generating grasps for the entire scene in a target-agnostic manner. Some methods directly extract scene-level semantic or geometric information. \cite{ni2020pointnet++,qin2020s4g} utilize PointNet++ \cite{qi2017pointnet++} to encode scene points and then directly regress grasp attributes. Recent methods in grasp generation predominantly adopt a two-stage approach: \textit{\textbf{where}} and \textit{\textbf{how}}. 

Chen et al. \cite{chen2023hggd} leverage ResNet \cite{he2016deep} to generate heatmaps, guiding the identification of grasping locations in image space. Similarly, works like \cite{fang2020graspnet,zhao2021regnet,wang2021graspness,liu2022transgrasp,liu2023joint} utilize PointNet++ as an encoder to extract global per-point features, assigning each point a ``graspness" or confidence score. Recognizing the importance of local information for effective grasping, these methods typically employ techniques like ball query or cylinder grouping to segment graspable regions around potential grasp centers. Chen et al. \cite{chen2023hggd} further augment this process by integrating points within the graspable regions with corresponding semantic features. A vanilla PointNet \cite{qi2017pointnet} is then used to extract both geometric and semantic features. In contrast, other studies \cite{zhao2021regnet,fang2020graspnet,wang2021graspness,wei2021gpr,liu2022transgrasp,liu2023joint} employ Multilayer Perceptrons (MLP) to encode deeper features based on the global per-point features extracted by PointNet++. Based on these extracted features, various mechanisms are developed to predict different grasp attributes tailoring to the specific grasp representation.

In contrast to the aforementioned scene-level methods that generate scene-wide grasps, we reformulate the 6-Dof grasp detection problem in a grasp-centric view. FlexLoG, our innovative framework, stands as the first to predict high-quality grasps solely based on local information. This focus on local data significantly improves the framework's ability to generalize to unseen objects, a benefit stemming from the object-agnostic characteristics of the regional points.

Besides, other methods such as Ten et al. \cite{ten2018using}, and Liang et al. \cite{liang2019pointnetgpd} follow a sample-evaluation strategy and also sample grasps locally. Compared to our streamlined approach, their reliance on complex, hand-crafted features is more time-consuming and results in lower grasp quality.  

\subsection{Target-Oriented Grasping}

% Few approaches focus on grasping specific objects or object parts in clutter.
Recently, several approaches \cite{murali20206,sundermeyer2021contact,liu2022ge} focus on grasping specific objects in cluttered by integrating an additional segmentation branch. Such a strategy, though practical, often results in excess computational load and cannot consistently yield high-quality grasps for the intended target. \cite{lu2023vl} utilizes a visual grounding algorithm for object detection, using bounding box filters to obtain object-level point clouds. These clouds are then processed by a network \cite{lu2022hybrid} trained on scene-level data, which can result in suboptimal grasping due to a domain mismatch with the training data.

Different from them, our framework, which is trained on regional grasp-centric data, can directly generate grasps on the target parts and seamlessly integrate various guidance methods, including object detection \cite{liu2023grounding, wang2023yolov7} and instance segmentation \cite{kirillov2023segment, ronneberger2015u}. This approach allows for direct processing of local point clouds rather than the whole scene, eliminating redundant computations. The object-agnostic nature of our method ensures higher grasp quality on target objects. Besides, our framework can readily incorporate part affordance concepts \cite{deng20213d, mo2021where2act} to facilitate grasping specific object parts.
 
\section{Problem Formulation}
\begin{figure}[b]
% \vspace{-0.8cm} % adjust distance above distance 
\centering
    {\includegraphics[width=7cm]{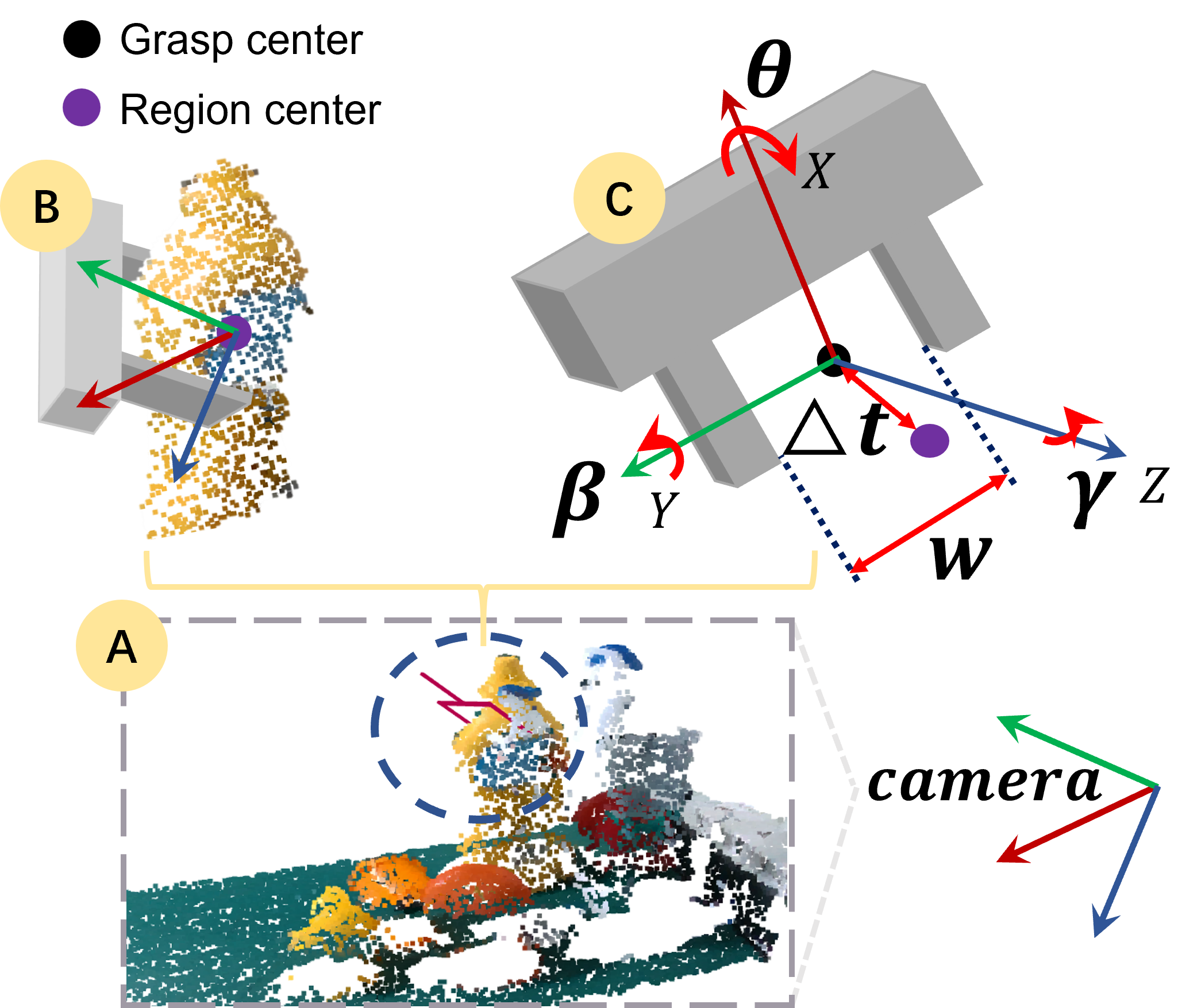}}
        \caption{A: The region is cropped from the scene point cloud in the camera frame. B: The local neighbor points are transformed to the local region frame. C: The regional grasp representation as $(\theta, \gamma, \beta, w, \Delta x, \Delta y, \Delta z)$.}
    \label{fig:representation} 
\end{figure}

Previous approaches generate scene-level grasps \cite{morrison2018closing,fang2020graspnet,chen2023hggd}, processing the whole scene information directly. Different from them, we reformulate this problem to region-level from a grasp-centric view. Assuming multiple grouped regions have been obtained from different guidance methods, we aim to predict grasps within each region, which can be formulated as
\begin{equation*}
    \boldsymbol{G_p} = \Phi(\mathcal{T}(f_{i} |i=1,...,K)),
\end{equation*}
where $f_i \in R^{N \times 3}$ is the $i$-th grouped region which is cropped from the scene point cloud and centered at $(x_p,y_p,z_p)$ in the camera frame, as shown in Fig. \ref{fig:representation}(A). $\mathcal{T}(\cdot)$ means canonical transformation (transform points to local region frame from camera frame in this paper). And $\Phi(\cdot)$ denotes the Local Grasp Model. To better fit the task  that predicts grasps within a region, one region-level grasp $\boldsymbol{g_p} \in \boldsymbol{G_p}$ is centered at $(x_p,y_p,z_p)$ and defined as:
\begin{equation*}
    \boldsymbol{g_p} = (\Delta x, \Delta y, \Delta z, \theta, \gamma, \beta,  w).
\end{equation*}
As Shown in Fig. \ref{fig:representation}(C), $(\theta,\gamma,\beta)\in [-\frac{\pi}{2}, \frac{\pi}{2}]$ are grasp Euler angles in the gripper frame and $w$ denotes the grasp width. Given that the guidance for grasp centers is often imprecise, there tends to be a misalignment between the actual grasp center and the regional center. To address this, we introduce an offset, $\Delta \boldsymbol{t}=(\Delta x, \Delta y, \Delta z)\in R^3$, which quantifies the deviation. Once $\boldsymbol{g_p}$ has been determined, the final grasp $\boldsymbol{g}$ can be represented as:
\begin{equation*}
    \boldsymbol{g} = (x_p+\Delta x, y_p+\Delta y, z_p+\Delta z, \theta, \gamma, \beta, w).
\end{equation*}

\begin{figure*}[t]
\centering
    \includegraphics[width=18cm]{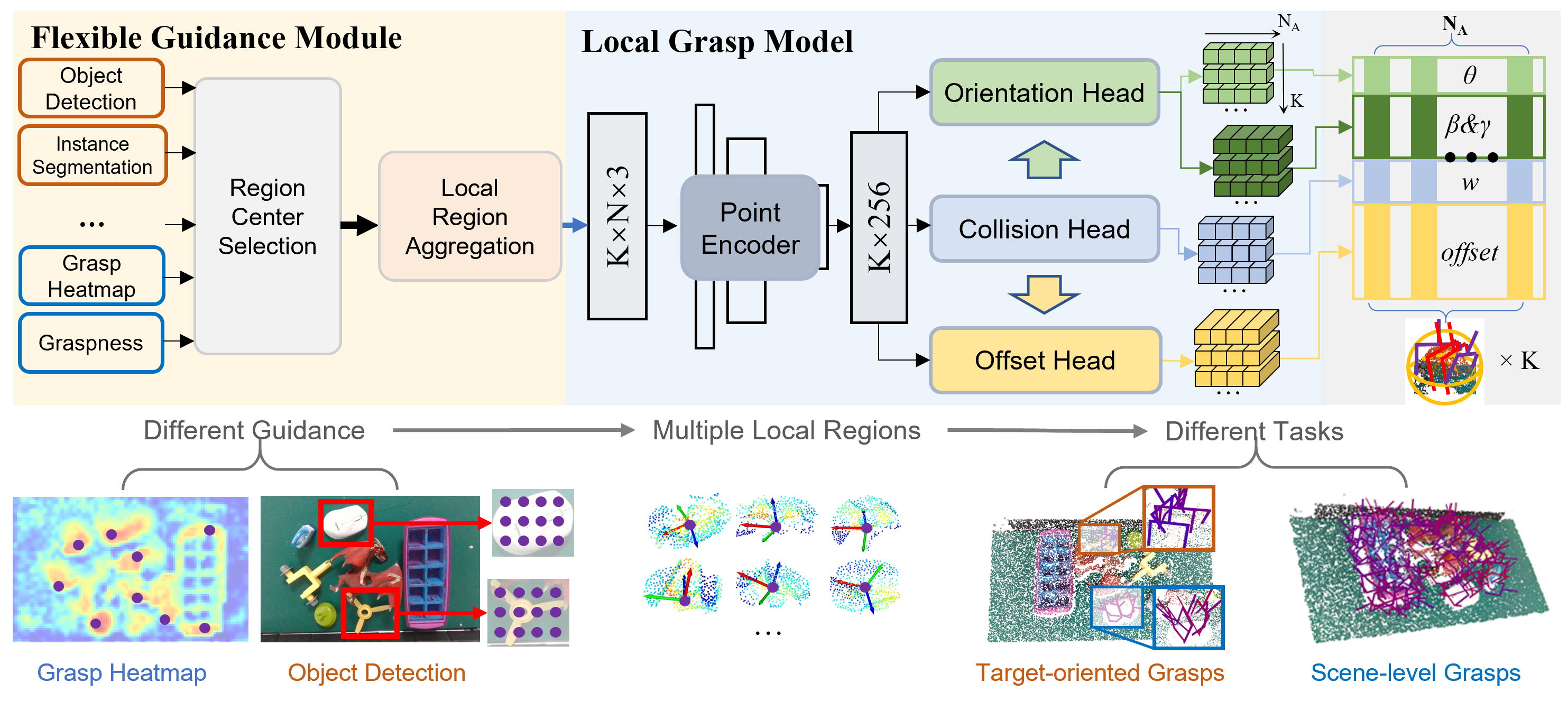}
    \caption{The architecture of FlexLoG. Taking a monocular observation image as input, the Flexible Guidance Module (FGM) utilizes different guidance methods (e.g., grasp heatmap for global guidance and object detection for local guidance) to identify potential graspable areas and sample points as regional centers. These points are then clustered into multiple local regions. The Local Grasp (LoG) Model then extracts geometric features and predicts grasps. Depending on the guidance method used in the FGM, the output is either scene-level or target-oriented grasps.}
    \label{fig:framework} 
% \vspace{-0.5cm} % adjust distance above distance 
\end{figure*}

\section{METHOD}
\subsection{Overview}

Our flexible 6-Dof grasp detection framework, \textbf{\textit{FlexLoG}}, consists of two main components: a \textit{\textbf{Flex}}ible Guidance Module (FGM) and a \textbf{\textit{Lo}}cal \textit{\textbf{G}}rasp Model (LoG). As is shown in Fig. \ref{fig:framework}, FGM leverages global or local guidance methods to sample potential grasp points. These points, designated as regional centers, are subsequently clustered with nearby points to form multiple distinct regions. LoG then extracts local geometric features from these regions and predicts grasps aligned with corresponding regional centers by employing three specially designed heads. The framework's region-level, grasp-centric methodology facilitates the integration of various guidance methods within the FGM and significantly enhances grasp quality, especially in unseen scenarios. LoG's proficiency in identifying grasp-related and object-agnostic geometric features from regional points is a critical factor in this enhanced performance.

\subsection{Flexible Guidance Module}

The \textit{\textbf{Flex}}ible Guidance Module (FGM) aims to identify and aggregate local regions with high graspability from the entire scene or specific targets. According to different downstream tasks, we categorize the guidance methods into two types: \textbf{\textit{scene-level}} and \textbf{\textit{target-oriented}} guidance. As depicted in Fig. \ref{fig:framework}, for \textbf{\textit{scene-level}} grasping, we mainly adopt two effective guidance methods, heatmap \cite{chen2023hggd} and graspness \cite{wang2021graspness}, to generate high-quality scene-level grasps. Both of them serve to indicate point-wise graspability and facilitate the sampling of points with high confidence, which are then used as regional centers for further aggregation.

Regarding \textbf{\textit{target-oriented}} grasping, local guidance methods such as object detection \cite{liu2023grounding} and semantic segmentation \cite{kirillov2023segment} can be utilized to sample regional centers from the predicted bounding boxes or segmentation masks. Furthermore, our FlexLoG framework is also compatible with other versatile guidance methods, such as part affordance \cite{deng20213d, mo2021where2act}, and even user-driven pixel selection through mouse clicks. 

To obtain regional points in the graspable areas, we employ the ball query method in conjunction with Furthest Point Sampling (FPS) \cite{qi2017pointnet}. This technique isolates a specified number of points within a spherical area from the scene's point cloud, effectively clustering them into local regions abundant with geometric features. As demonstrated in Fig. \ref{fig:representation}, we then transform these points ${f_{i}} \in R^{N \times 3}, i=1,...,K$, from the camera frame to the local region frame, resulting in a set of regional points ${\mathcal{T}(f_{i} |i=1,...,K)}$, where $K$ denotes the number of regions (i.e., sampled regional centers).

\begin{figure}[t]
% \vspace{-0.8cm} % adjust distance above distance 
\centering
    {\includegraphics[width=8.5cm]{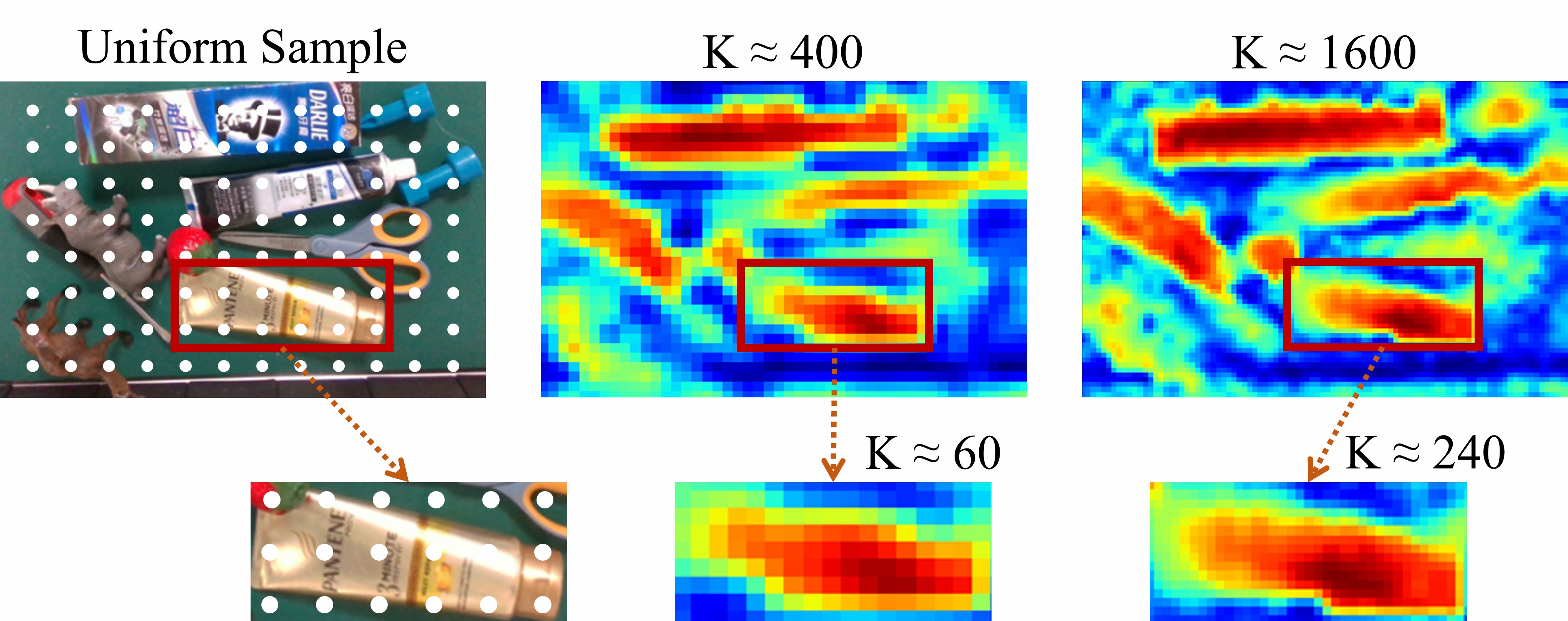}}
        \caption{Local grasp scores can be spliced to form a grasp heatmap, illustrating the graspability. As the number of sampled centers $K$ increases, so does the heatmap's resolution, leading to a more accurate depiction of the graspable areas.}
    \label{fig:gridsample}
\end{figure}

Addressing the challenge of generating scene-level grasps without external guidance, we employ a heuristic approach to analyze graspable areas in the scene. As shown in Fig. \ref{fig:gridsample}, in this scenario, we treat the entire scene as the whole graspable area, which is actually a special case of global guidance. We start by uniformly sampling pixels in a 2D mesh grid across the scene with a fixed grid size. Each grid center correlates to a specific point in 3D space, which is then utilized as a center for local region aggregation. Based on the aggregated regions, LoG can generate local grasps efficiently. Subsequently, the highest grasp scores within these regions are spliced to form a grasp heatmap, serving as a visual representation of the potential distribution of graspable areas. Notably, the mesh grid size is adjustable, allowing for a tailored balance between the quality of grasp prediction and time efficiency. 

\subsection{Local Grasp Model} \label{section_LoG}

From a grasp-centric view, each grasp is fundamentally influenced by the information of its surrounding region. To this end, we develop the \textbf{\textit{Lo}}cal \textit{\textbf{G}}rasp Model (LoG), a robust network designed for extracting local geometric features and predicting multiple regional grasps. While existing studies \cite{zhao2021regnet, fang2020graspnet, wang2021graspness, liu2022transgrasp} utilize feature extractors for local geometry, they predominantly rely on global features from PointNet++. The more recent HGGD \cite{chen2023hggd} is more similar to our proposed framework but still requires necessary pixel-wise grasp attribute prediction from global images. Our LoG, however, is the first model capable of predicting high-quality grasps solely utilizing local information, represented as $\boldsymbol{g_p} = (\theta, \gamma, \beta, w, \Delta x, \Delta y, \Delta z)$. This advanced capability aligns seamlessly with the functionalities of our FGM.
% In contrast, our method is the first framework capable of predicting high-quality grasps $\boldsymbol{g_p} = (\theta, \gamma, \beta, w, \Delta x, \Delta y, \Delta z)$ exclusively using local information, marking an innovative advancement in the field.

\begin{figure}[t]
% \vspace{-0.8cm} % adjust distance above distance 
\centering
    {\includegraphics[width=6cm]{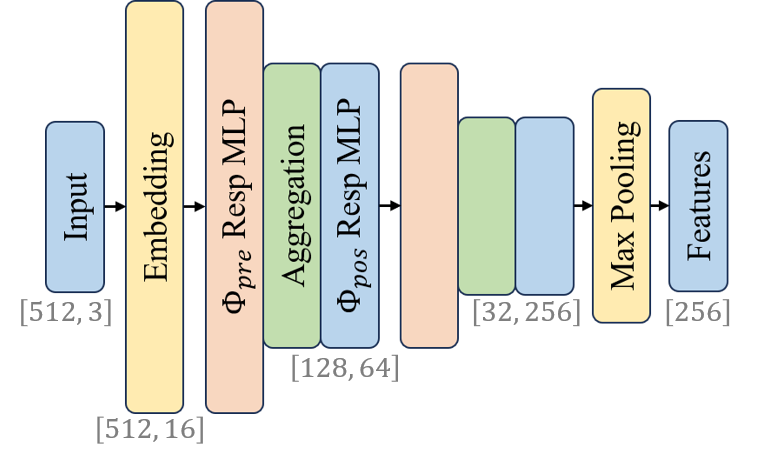}}
        \caption{The proposed light-weighted PointMLP-based encoder structure of Local Grasp Model.}
    \label{fig:pointmlp}
\end{figure}

LoG processes the transformed local points ${\mathcal{T}(f_{i} |i=1,...,K)}$ and predicts grasps for each region. As illustrated in Fig. \ref{fig:pointmlp}, our proposed point encoder, inspired by PointMLP \cite{ma2022rethinking}, is adept at local geometric feature extraction and achieves a better balance between efficiency and precision. With the local features extracted, LoG employs three specialized heads – \textbf{\textit{Collision Head}}, \textbf{\textit{Orientation Head}}, and \textbf{\textit{Offset Head}} – all based on Multilayer Perceptron (MLP) architecture, to predict various grasp attributes: $w$, $(\theta,\beta,\gamma)$, and $(\Delta x, \Delta y, \Delta z)$. These attributes collectively form the 6-Dof grasps $\{\boldsymbol{g_p}_{(i,j)}|i=1,...,K, j=1,...,N_A\}$ within each region, where $N_A$ denotes the number of grasp anchors. 

In the \textbf{\textit{Collision Head}} of LoG, informed by a predefined range, grasp width prediction is approached as a regression problem to avoid collisions with nearby objects. These predicted widths, encapsulating vital local collision information, are subsequently utilized in the Offset and Orientation Heads. For the \textbf{\textit{Offset Head}}, our approach mirrors that of HGGD, regressing the normalized grasp location offset along three axes, thereby refining grasp placements within each region. In the \textbf{\textit{Orientation Head}}, we first predict the grasp in-plane rotation angle $\theta$, a critical step due to its role as a precursor for spatial rotation angles $(\beta, \gamma)$ in our Euler angle setting. The angle range $\theta \in [-\frac{\pi}{2}, \frac{\pi}{2}]$ is discretized into $k_{\theta}=6$ bins. We employ a combined approach of bin classification and residual regression within each bin to derive the final $\theta$ value. Furthermore, the predicted $\theta$ values are incorporated as contextual data for $(\beta, \gamma)$ predictions. Following the non-uniform anchor sampling strategy \cite{chen2023hggd}, our strategy views spatial rotation prediction as a multi-label classification task. With $7$ anchors each for $(\beta, \gamma)$, we generate up to $N_A=49$ possible grasps per region and preserve those with the highest scores. 

Compared with scene-level methods \cite{zhao2021regnet,fang2020graspnet,wang2021graspness,liu2022transgrasp,chen2023hggd}, our LoG exclusively focuses on regional data and local geometric feature extraction, which significantly enhances grasp detection quality. A vital aspect of our strategy involves the adoption of regional point clouds, which frequently lack complete object shape information, as the network input. This compels the network to adapt to learning object-agnostic features, consequently leading to significantly enhanced generalization, especially in unseen scenarios.
% our LoG is totally based on region-level data and focuses on the local geometric feature excavation, bringing a higher grasp quality. Importantly, most of our training data does not contain a full object shape, which forces the network to learn the object-agnostic features, showing a much better generalization to unseen objects.

\subsection{Regional Grasp-centric Dataset Generation} \label{section_data}

To effectively train the Local Grasp Model, we generate a new dataset containing regional point clouds and local grasp labels derived from existing scene-level annotations in \cite{fang2020graspnet}. The dataset creation involves selecting potential grasp centers and cropping local regions around them. Intuitively, sampling grasp center candidates randomly across the entire space is straightforward but proved inefficient, yielding a high proportion of invalid data. Alternatively, sampling centers directly from grasp label projections, as suggested in \cite{chen2023hggd}, leads to suboptimal results during inference due to domain shift caused by heatmap prediction deviation.

\begin{figure}[t]
% \vspace{-0.8cm} % adjust distance above distance 
\centering
    {\includegraphics[width=9cm]{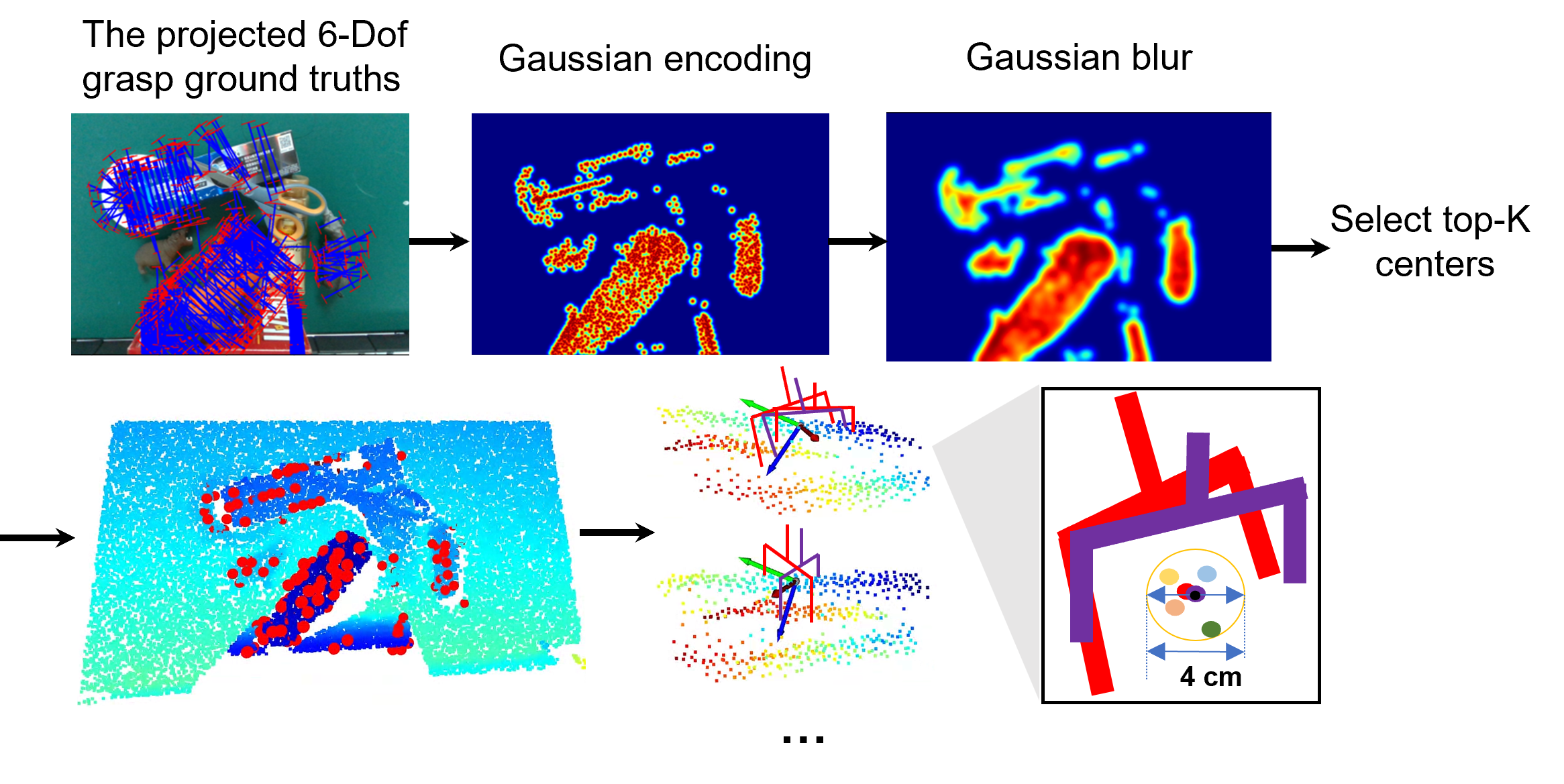}}
        \caption{The pipeline of region-level grasp-centric data generation. Grasp centers are sampled using the Gaussian-based strategy. Then, local neighbor points around each center are cropped as regions. Only the grasp labels within a radius of 2 cm from the region center are preserved.}
    \label{fig:datageneration}
\end{figure}

Addressing these challenges, we implement a Gaussian-based strategy for center selection, which introduces a small proportion of invalid data, naturally incorporating noise into the training process. As shown in Fig. \ref{fig:datageneration}, this process begins by projecting 6-Dof grasp ground truths to 4-Dof planar grasps. Each planar grasp center is encoded using a Gaussian kernel. Subsequently, we apply another Gaussian filter to create a blurred heatmap, depicting the distribution of graspable areas. Centers are then selected from this heatmap through grid-based sampling and converted into 3D points using corresponding depth values. For each center, a sphere with a radius between 6 and 12 cm – informed by the gripper width – is defined, and the ball query method \cite{qi2017pointnet} is applied to extract relevant points.

Label processing involves preserving only those grasp labels within a 2 cm Euclidean distance from the corresponding region centers, reinforcing the grasp-centric nature of our framework and dataset. This criterion ensures that LoG focuses on generating grasps near regional centers, which is critical for target-oriented grasping. Points and labels are then transformed from the camera frame to the local frame, normalizing the data distribution.

Through the above steps, we construct a substantial dataset of millions of grasp-centric and object-agnostic regional data, providing a basis for the LoG training.

\subsection{Implementation Details} \label{implementation}

As depicted in Section. \ref{section_LoG}, our FlexLoG incorporates classification and regression components. Thus, the overall loss, which is equal to the regional grasp prediction loss $L$, could then be formulated with different loss terms as
\begin{equation*}
    L = a \times (L_\theta^{cls} + L_\theta^{reg}) + b \times L_{w} + c \times L_{offset} + d \times L_{anchor},
\end{equation*}
where $L_\theta^{cls}$ represents the cross-entropy loss for the bin classification of $\theta$. A Smooth $L1$ loss $L_\theta^{reg}$ is adopted for the residual regression within one bin. Another two Smooth $L1$ loss $L_{w}$ and $L_{offset}$ are employed to supervise the regression of local grasp width and center offset. $L_{anchor}$ represents the multi-label classification loss calculated using focal loss \cite{lin2017focal} for the non-uniform anchor sampling strategy.

For FGM, we directly adopt the pretrained heatmap checkpoint \cite{chen2023hggd} and the reimplemented graspness model \cite{wang2021graspness} as our scene-level guidance. With specific guidance, we aggregate $K = 48$ local regions by default. As for the uniform sampling without guidance, we uniformly sample points in the 2D mesh grid with grid size $12$ pixels by default. However, in addition to the default sampling settings, considering the trade-offs between inference time and detection performance, it is practicable to adjust the center number of local regions in different scenarios.

For LoG, similar to the local region settings in \cite{chen2023hggd}, we aggregate local point clouds with  $N = 512$ points via Furthest Point Sampling (FPS) for each region. Notably, the same anchor shifting strategy is employed to generate more accurate grasps. 

\section{EXPERIMENT}
To thoroughly evaluate the performance of our proposed framework, we conduct experiments both on the dataset and the real robot platform.

\begin{table*}[t]
\small
\renewcommand{\arraystretch}{1.3}
\caption{Results on GraspNet Dataset, showing APs on RealSense/Kinect split and method efficiency}
\begin{center}
\begin{tabular}{c|c c c c c} 
\hline
\textbf{Method}&{\textbf{Seen}} $\uparrow$ &{\textbf{Similar}} $\uparrow$ &{\textbf{Novel}} $\uparrow$ &{\textbf{Average}} $\uparrow$&{\textbf{FPS}} $\uparrow$\\
\hline
GPD \cite{ten2017grasp} & 22.87 / 24.38 & 21.33 / 23.18 & 8.24 / 9.58 & 17.48 / 19.05 & -\\
RGB Matters \cite{gou2021rgb} & 27.98 / 32.08 & 27.23 / 30.40 & 12.25 / 13.08 & 22.49 / 25.19 & 2.3\\
REGNet \cite{zhao2021regnet} & 37.00 / 37.76 & 27.73 / 28.69 & 10.35 / 10.86 & 25.03 / 25.77 & 2.2\\
TransGrasp \cite{liu2022transgrasp} & 39.81 / 35.97 & 29.32 / 29.71 & 13.83 / 11.41 & 27.65 / 25.70 & -\\
Graspness \cite{wang2021graspness} & 67.12 / 63.50 & 54.81 / 49.18 & 24.31 / 19.78 & 48.75 / 44.15 & $\sim$10\\
Scale Balanced Grasp \cite{ma2023towards} & 63.83 / \quad-\quad\quad & 58.46 / \quad-\quad\quad & 24.63 / \quad-\quad\quad & 48.97 / \quad-\quad\quad & -\\
HGGD \cite{chen2023hggd} & 59.36 / 60.26 & 51.20 / 48.59 & 22.17 / 18.43 & 44.24 / 42.43 & \textbf{28}\\
\hline
FlexLoG (\textit{Uniform Sampling}) & 68.45 / \underline{64.05} & 61.61 / \underline{55.72} & 27.94 / 21.59 & 52.67 / \underline{47.12} & 10\\ 
FlexLoG (\textit{Graspness Guidance}) & \underline{69.23} / 62.86 & \underline{62.40} / 53.22 & \underline{29.63} / \textbf{25.15} & \underline{53.75} / 47.08 & 9.0\\ 
FlexLoG (\textit{Heatmap Guidance}) & \textbf{72.81} / \textbf{69.44} & \textbf{65.21} / \textbf{59.01} & \textbf{30.04} / \underline{23.67} & \textbf{56.02} / \textbf{50.67} & \underline{26}\\ 
\hline
\end{tabular}
\label{graspnet}
\end{center}
\end{table*}

% \begin{table}[t]
% \small
% \renewcommand{\arraystretch}{1.3}
% \caption{Guidance Ablation Experiments, showing APs on RealSense/Kinect split and method efficiency}
% \begin{center}
% \begin{tabular}{c c c c} 
% \hline
% {\textbf{Method}}&{\textbf{Average}} $\uparrow$ &{\textbf{FPS}} $\uparrow$\\
% \hline
% Uniform Sampling & 52.67 / 47.12 & 10 \\ 
% \hline
% Objectness Guidance
% & 52.40 / 45.69 & 7.6 \\
% Graspness Guidance\cite{wang2021graspness} & \textbf{56.27} / \underline{49.38} & 7.2 \\
% Heatmap Guidance\cite{chen2023hggd} & \underline{56.02} / \textbf{50.67} & \textbf{26} \\
% \hline
% \end{tabular}
% \\ \quad
% \label{ablation_model2}
% \end{center}
% \end{table}

\subsection{Performance Evaluation}

\subsubsection{\textbf{Scene-Level Grasping}}
To better compare overall performance with other methods, we firstly evaluate them in the scene-level grasping situation. GraspNet-1Billion dataset \cite{fang2020graspnet} is widely used in grasp detection as a standard evaluation platform for the task of general robotic grasping, containing RGBD images captured in the real world from 190 cluttered scenes and more than 1 billion grasp annotations. We utilize the method in Section. \ref{section_data} to generate the regional grasp-centric data based on the GraspNet-1Billion dataset. Around \textbf{6.5M} local regions are obtained from the training split to train the proposed LoG and compare its performance with other methods. 

Following previous works, the \textbf{Average Precision (AP)} \cite{fang2020graspnet} evaluation metric is adopted, which adopts the same average precision calculation metric in object detection task for the force-closure scores \cite{qin2020s4g} of the top 50 grasps after non-maximum suppression.

\begin{figure}[t]
% \vspace{-0.8cm} % adjust distance above distance 
\centering
    {\includegraphics[width=8cm]{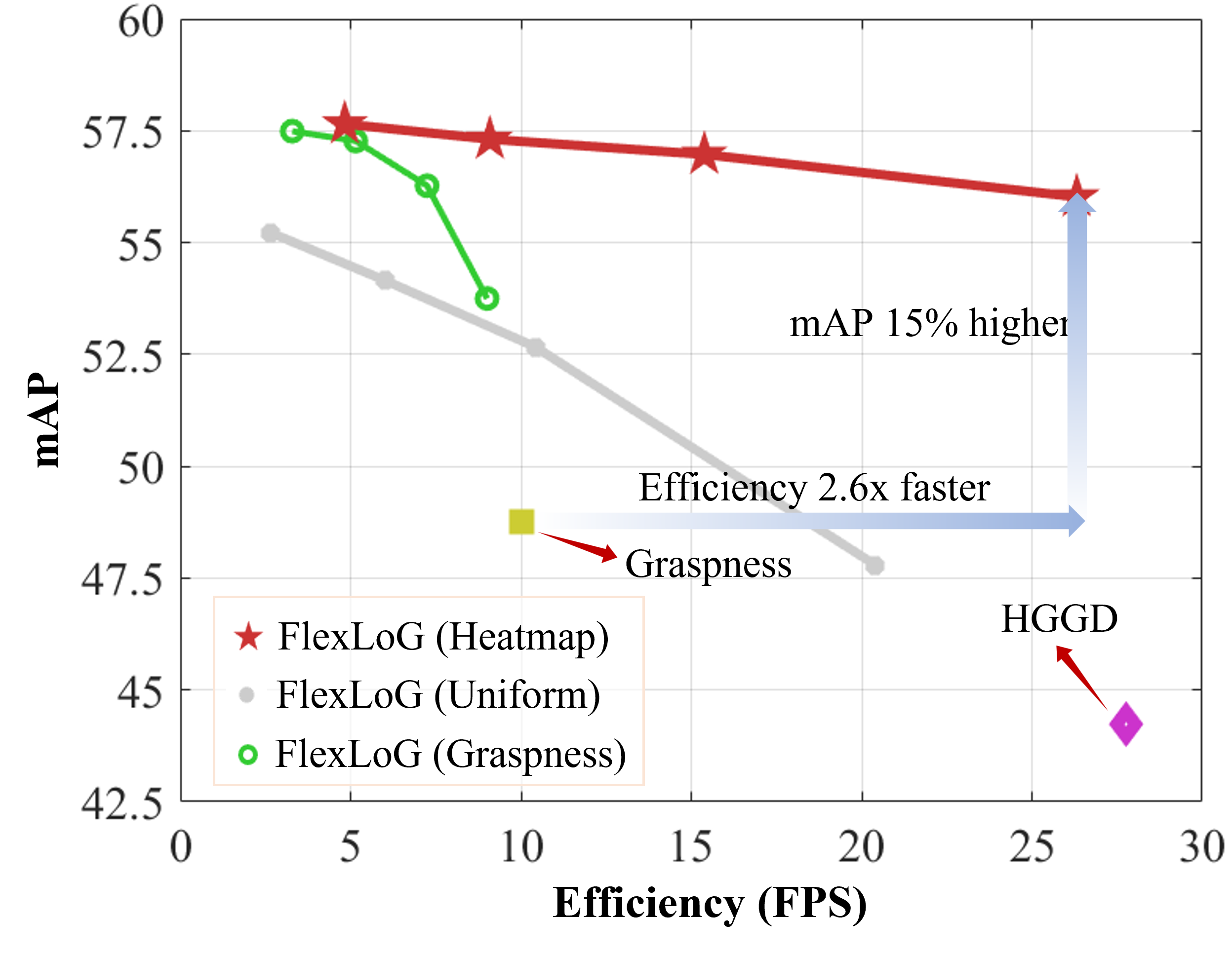}}
        \caption{The performance curve (mAP on all test scenes) on GraspNet dataset. Our framework with heatmap guidance outperforms previous state-of-the-art method Graspness with 2.6 times faster speed and 15\% higher mAP.}
    \label{fig:line}
\end{figure}

For the FGM, we adopt uniform sampling (without any guidance), point-wise graspness guidance, and heatmap guidance to aggregate local regions and conduct scene-level grasp detection. As illustrated in Table \ref{graspnet}, FlexLoG makes significant gains and achieves the new state-of-the-art on the GraspNet-1Billion benchmark. It is worth noting that our heatmap-guided approach achieves \textbf{10.4/9.83} and \textbf{5.73/3.89} performance gains on the similar and novel splits compared to the previous state-of-the-art method Graspness, which demonstrates that our region-level and grasp-centric method has a more robust generalization to unseen scenarios. Furthermore, FlexLoG based on local region grasp generation is robust to different guidance inputs and manages to achieve high-quality grasp detection results even without any guidance, which indicates FlexLoG's potential for other target-oriented grasp scenarios.

Besides, we draw curves of (inference efficiency (FPS) - mAP) by sampling different numbers of candidate region centers to conduct detailed analysis and evaluation. With more candidate centers, methods detect grasps with higher mAPs and object coverage rates but at a slower speed. As shown in Fig. \ref{fig:line}, our method with heatmap guidance achieves state-of-the-art grasp detection results at different speeds (center numbers). Even without any guidance, our method can still outperform other methods by a large margin. Thus, the experiment results prove that FlexLoG is a flexible framework for high-quality and real-time grasping.

\subsubsection{\textbf{Target-Oriented Grasping}}
To further evaluate the performance of our LoG for target-oriented grasping scenarios, we modified the evaluation metric from \textbf{AP} to \textbf{T}arget-\textbf{O}riented \textbf{A}verage \textbf{P}recision (\textbf{TOAP}) and adjusted the evaluation process following the same procedure of TOGNet\cite{xie2024target}. 

Specifically, for fair comparisons, we use the ground truth object segmentation mask from the GraspNet-1Billion dataset to randomly select one object, not fully occluded by others, as the target for each RGB-D image.  Since there are 512 RGB-D images with different cameras and viewpoints for each scene, and approximately 10 objects per scene, all objects can be selected as targets. During evaluation, we compute the distance from the detected grasp centers to the target object mesh model, retaining only the grasps on the target. Finally, due to reduced grasp diversity, we compute force-closure scores \cite{qin2020s4g} for the \textbf{top 10} grasps after non-maximum suppression and use scores under different friction coefficients to derive \textbf{TOAP}.

\begin{table*}[t]
\small
\renewcommand{\arraystretch}{1.25}
\caption{Target-oriented APs on GraspNet Dataset (RealSense/Kinect)}
\vspace{-0.2cm}
\centering
\begin{threeparttable}
\begin{tabular}{c|>{\centering\arraybackslash}p{2.15cm}>{\centering\arraybackslash}p{2.15cm}>{\centering\arraybackslash}p{2.15cm}|>{\centering\arraybackslash}p{2.15cm}} 
\hline
\textbf{Method}&{\textbf{Seen}}&{\textbf{Similar}}&{\textbf{Novel}} &{\textbf{Mean}}\\
\hline
GraspNet-baseline \cite{fang2020graspnet} & 22.64 / 13.28 & 20.63 / 12.67 & 8.35 / 3.85 & 17.21 / \;9.93\;\\
Scale Balanced Grasp \cite{ma2023towards} & 38.04 / \quad-\quad\quad\ & 33.94 / \quad-\quad\quad\ & 15.97 / \quad-\quad\quad\ & 29.32 / \quad-\quad\quad\ \\
HGGD \cite{chen2023hggd} & 38.91 / 34.82 & 34.70 / 28.77 & 16.73 / 12.02 & 30.11 / 25.20\\
Graspness \cite{wang2021graspness} & 44.80 / 34.81 & 37.67 / 29.61 & 18.06 / 12.82 & 33.51 / 25.75\\
TOGNet \cite{xie2024target} & \textbf{51.84} / \textbf{49.60} & \textbf{46.62} / \textbf{40.03} & \textbf{23.74} / \textbf{19.58} & \textbf{40.73} / \textbf{36.40}\\
\hline
LoG  & 50.57 / 44.67 & 44.59 / 39.37 & 22.59 / 16.04 & 39.25 / 33.36  \\
\hline
\end{tabular}
``-'': Result unavailable
\label{graspnet_toap}
\end{threeparttable}
\vspace{-0.2cm}
\end{table*}

The evaluation results are shown in Table \ref{graspnet_toap}. Notably, our LoG model achieves significantly higher results than other baseline methods and delivers comparable target-oriented grasping performance to TOGNet. However, unlike TOGNet, which utilizes both RGB and XYZ features, our LoG model uses only local XYZ features, making it more convenient for certain applications.

\begin{table}[t]
\small
\renewcommand{\arraystretch}{1.3}
\caption{Model Ablation Experiments, showing APs on RealSense/Kinect Split and method efficiency}
\begin{center}
\begin{tabular}{c c c} 
\hline
\textbf{Method}&{\textbf{Average} $\uparrow$}&{\textbf{FPS} $\uparrow$}\\
\hline
Ours & \underline{56.02} / 50.67 & \underline{26} \\
\hline
PointNet & 52.65 / 47.98 & \textbf{30} \\
Wider PointMLP & \textbf{56.32} / \textbf{51.86} & 21 \\
Deeper PointMLP & 54.48 / \underline{51.02} & 20 \\
\hline
\end{tabular}
\\ \quad
\label{ablation_model1}
\end{center}
\end{table}

\subsubsection{\textbf{Ablation Studies}}

To further analyze the role of the proposed LoG, we design ablation studies on the local grasp model backbone. We mainly discuss the performance of the point encoder with different architectures. As illustrated in Table \ref{ablation_model1}, firstly, we change our encoder to PointNet \cite{qi2017pointnet} as the same as \cite{chen2023hggd}, which shows a faster speed but an unignorable performance drop. Then, we explore the influence of the PointMLP structure. \textit{Wider PointMLP} means double the network width (embedding dimension), and \textit{Deeper PointMLP} means double the network depth (network grouping and MLP layer number). Experiments show that a wider network slightly improves grasp detection performance but with a much slower speed. Moreover, a deeper and slower network cannot bring any increase in model performance. In conclusion, the overall results show the efficiency of our designed light-weighted PointMLP encoder for local grasp detection.

\subsection{Real Robot Experiments}

\begin{figure}[t]
% \vspace{-0.8cm} % adjust distance above distance 
\centering
    {\includegraphics[width=8cm]{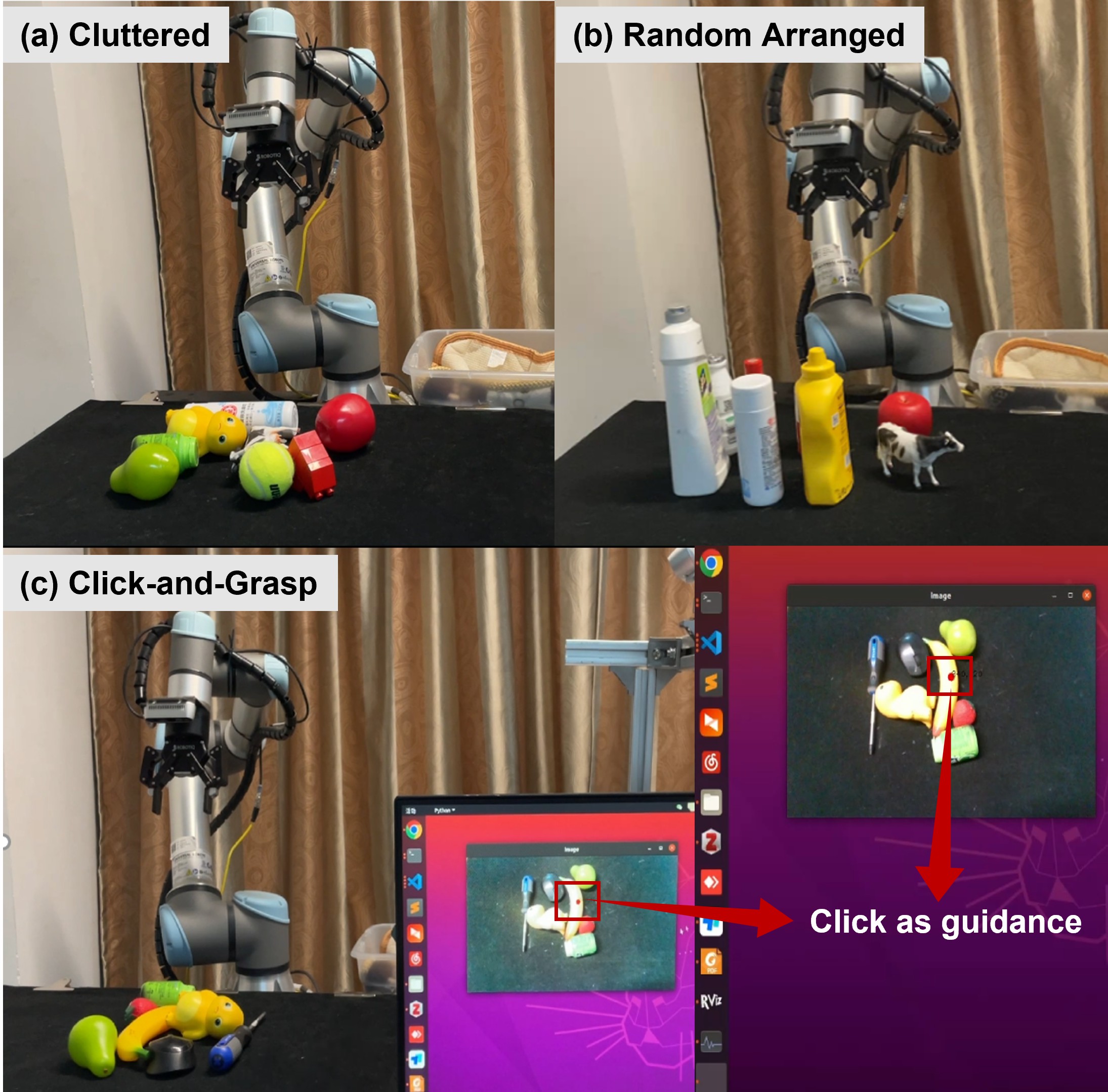}}
        \caption{Three settings of real-world robot experiments. (a) \textbf{Cluttered}: the poses of objects are randomized by shaking them in a box. (b) \textbf{Random Arranged}: arrange objects randomly and simulate the state of them in daily life. (c) \textbf{Click-and-Grasp}: the points of click are used as local guidance for target-oriented grasping.}
    \label{fig:real}
\end{figure}

We also perform real-world robot grasping experiments using a UR-5e robot equipped with a Robotiq 2-finger parallel-jaw gripper. To capture single-view RGBD images, we employ a Realsense-D435i camera. Our experiment procedure follows that of previous studies, but we quantitatively evaluate our algorithms in three different settings: \textbf{\textit{Cluttered}}, \textbf{\textit{Random Arranged}}, and \textbf{\textit{Click-and-Grasp}}. We assemble a collection of 30 objects with diverse shapes and sizes commonly encountered in everyday life, 15 of which have never been seen in the GraspNet-1Billion dataset. Subsequently, we randomly select and arrange 7 to 9 of these objects in each scene, placing them on the table in various orientations. 

As illustrated in Fig. \ref{fig:real}, to evaluate FlexLoG's performance of \textit{scene-level} grasping, we design the \textbf{\textit{Cluttered}} and \textbf{\textit{Random Arrange settings}}. To evaluate the performance of \textit{target-oriented} grasping, we demonstrate a \textbf{\textit{Click-and-Grasp}} setting for simulating local guidance. 

Specifically, for the \textbf{\textit{Cluttered}} setting, following \cite{liang2019pointnetgpd}, we shake a box with all the objects in it, ensuring that the poses of all the objects are randomized. For the \textbf{\textit{Random Arranged}} setting, we consider the general states of objects in daily life, such as a bottle standing upright. Then we arrange and organize these objects randomly on a tabletop. For the \textbf{\textit{Click-and-Grasp}} setting, to evaluate the target-oriented grasping performance of LoG, we randomly select one object, click at a specific prehensile part, and use the points in the local region to generate grasps. Please see the supplementary video for the demonstration. We adopt the \textbf{Success Rate} and \textbf{Completion Rate} to evaluate the performance. Note that for the Click-and-Grasp setting, success is contingent upon successfully grasping the selected object.

It is worth noting that, in the \textbf{\textit{Cluttered}} and \textbf{\textit{Random Arranged}} settings, we employ a naive uniform sampling strategy instead of any other global guidance models. Table \ref{tab:real} reports that our method achieves an average grasp success rate of 95\% across 18 test scenarios, with a 100\% scene completion rate. It indicates that the FlexLoG can generalize to the real world and generate high-quality grasps efficiently. Some failures are observed in situations where objects have smooth surfaces (e.g., the bottle) and the gripper slides over the object, or sometimes the gripper collides with other objects in the clutter.

\begin{table}[t]
\renewcommand{\arraystretch}{1.25}
\caption{Results of robotics experiments}
\begin{center}
\begin{threeparttable}
\begin{tabular}{c|c c c c} 
\hline
\textbf{Scene}&\textbf{Success Rate}$^{1}$ &\textbf{Completion Rate}$^{2}$ \\
\hline
Cluttered & 44 / 46 = 96 \% & 6 / 6 = 100 \% \\
Random Arranged & 48 / 51 = 94 \% & 6 / 6 = 100 \% \\
Click-and-Grasp & 48 / 51 = 94 \% & 6 / 6 = 100 \% \\
\hline
Total & 140 / 148 = \textbf{95 \%} & 18 / 18 = \textbf{100 \%} \\
\hline
\end{tabular}
$^1$ \textbf{Success} / \textbf{Attempt} = \textbf{Success Rate} \\
$^2$ \textbf{Cleared Scene} / \textbf{Total Scene} = \textbf{Completion Rate}
\end{threeparttable}
\label{tab:real}
\end{center}
\end{table}

\section{CONCLUSION}

In this paper, we rethink the 6-Dof grasp detection problem and propose a flexible 6-Dof grasp framework, FlexLoG. Through a novel grasp-centric view, the designed Local Grasp Model can be integrated with either global or local guidance methods for scene-level grasping or target-oriented grasping. Our framework achieves state-of-the-art performance on GraspNet-1Billion Dataset. Besides, the quantitative real-world robot grasping experiments demonstrate the effectiveness of our method. However, the LoG uses only point cloud, which is prone to generate low-quality grasps when the point cloud is of poor quality (e.g., transparent or reflective objects). In the future, we consider adding more semantic information to enhance robustness.

\bibliography{reference}

\addtolength{\textheight}{-12cm}   % This command serves to balance the column lengths

\end{document}